\documentclass{article} 
\usepackage{iclr2020_conference,times}

\usepackage{amsmath,amsfonts,bm}

\def\eqref#1{equation~\ref{#1}}

\def\1{\bm{1}}

\DeclareMathAlphabet{\mathsfit}{\encodingdefault}{\sfdefault}{m}{sl}
\SetMathAlphabet{\mathsfit}{bold}{\encodingdefault}{\sfdefault}{bx}{n}

\usepackage{hyperref}
\usepackage{url}

\title{Tigrinya Neural Machine Translation with Transfer Learning for Humanitarian Response}

\author{Alp \"Oktem, Mirko Plitt \& Grace Tang \\
Translators without Borders\\
30 Main Street STE 500\\
Danbury, CT 06810, USA \\
\texttt{\{alp,mirko,grace\}@translatorswithoutborders.org} \\
}

\iclrfinalcopy 
\begin{document}

\maketitle
\setcounter{footnote}{-1}
\begin{abstract}
We report our experiments in building a domain-specific Tigrinya-to-English neural machine translation system. We use transfer learning from other Ge'ez script languages and report an improvement of 1.3 BLEU points over a classic neural baseline. We publish our development pipeline as an open-source library and also provide a demonstration application.
\end{abstract}

\section{Introduction}

Tigrinya (also spelled Tigrigna) is an Ethiopic language spoken by around 7.9 million people in Eritrea and Ethiopia. It is neither supported by any commercial machine translation (MT) provider, nor has any publicly available models. Refugees who speak Tigrinya face language and communication barriers when arriving in Europe. An MT system could improve access to  information and enable two-way communication so refugees have a voice and can share their needs.

The complex morphological structure of Tigrinya makes it especially challenging for statistical MT \citep{Tedla2016TheEO, teferra-abate-etal-2018-parallel}. Neural MT, on the other hand, can overcome these problems with methods like subword segmentation \citep{sennrich-etal-2016-neural} and lead to more accurate models \citep{kalchbrenner-blunsom-2013-recurrent-continuous, DBLP:journals/corr/BahdanauCB14}. Known to be a data-hungry technology, it is now possible to train neural-based MT for Tigrinya thanks to recently released public datasets \citep{agic-vulic-2019-jw300, teferra-abate-etal-2018-parallel}. An advantage of neural MT is the availability of techniques like cross-lingual transfer learning \citep{zoph-etal-2016-transfer} and multilingual training \citep{dong-etal-2015-multi} which help leverage data from other languages and are especially suitable in low-resource scenarios \citep{neubig-hu-2018-rapid}.

In this paper, we explain the development of a Tigrinya-to-English neural MT model using publicly available datasets in Ge'ez-scripted languages. Our models are further adapted to the humanitarian domain to improve the translation capabilities of Translators without Borders (TWB), a non-profit organization offering language and translation support for humanitarian and development agencies, and other non-profit organizations.

\vspace*{-1mm}

\section{Experiments}
\vspace*{-1mm}
\subsection{Data}

We gathered an internal dataset from sentences in TWB's translation memories. This dataset is both used for in-domain training and for testing. Two hundred sentences of varying lengths were selected randomly as a test set. This and other sources of public parallel corpora used in this work are listed in Table \ref{data-table}.

\begin{table}[t]
\caption[caption]{Parallel corpora used in this work. Dataset names marked with an asterisk are available through OPUS repository \citep{TIEDEMANN12.463, Christodouloupoulos2015}. Ethiopian languages corpus \citep{teferra-abate-etal-2018-parallel} is also openly available online\footnotemark.}
\label{data-table}
\begin{center}
\begin{tabular}{l|cccccccc}
  & JW300* & Ethiopian & Bible- & Global & GNOME* & Tanzil* & TWB & \textbf{TOTAL}\\
  &  & corpus & uedin* & Voices* &  &  & 
\\ \hline \\
Amharic  & 722K  & 66K               & 61K         & 1,6K          & 57K   & 94K    & - & 1M   \\
Ge'ez    & -     & 11K               & -           & -             & -     & -      & -   & 11K  \\
Tigrinya & 400K  & 36K               & -           & -             & -     & -      & 2.5K & 439K \\
\end{tabular}
\end{center}
\end{table}

\vspace*{-2mm}
\subsection{Experimental setup}

The transfer-learning-based training process consists of three stages. First, we train the model on a shuffled mix of all datasets totaling up to 1.45 million sentences. Second, we fine-tune the model on Tigrinya using only the Tigrinya portion of the mix (438,000 sentences). In the third phase, we fine-tune on the training partition of our in-house data (2,300 sentences). As a baseline, we skip the first multilingual training step and use only Tigrinya data. The model is later fine-tuned to in-domain data in the same way.

OpenNMT-py toolkit \citep{klein-etal-2018-opennmt} is used for training the models. The model consists of an 8-head Transformer \citep{DBLP:conf/nips/VaswaniSPUJGKP17} with 6-layer hidden units of 512 unit size. A token-batch size of 4096, 2048 and 10 was selected for multilingual, unilingual and in-domain training respectively. As for the optimizer, Adam \citep{Kingma2014AdamAM} was chosen with 4000 warm-up steps. Trainings were performed until no further improvement was recorded in development set perplexity in the last 5 validations. This resulted in 73,500, 85,000 and 85,240 steps for each stage.

\footnotetext{\url{http://github.com/AAUThematic4LT/Parallel-Corpora-for-Ethiopian-Languages}}

Byte-pair encoding (BPE) models were trained separately for Latin script and Ge'ez script using 6,000 steps. English sentences were lowercased and tokenized beforehand using Moses tokenizer \citep{koehn-etal-2007-moses}. Ge'ez-scripted sentences were tokenized using a punctuation separation script\footnote{\url{http://github.com/translatorswb/mt-tools}}.

\vspace*{-2mm}

\subsection{Results}

We report our test set scores at each stage together with the baseline using various commonly used automatic evaluation metrics in Table \ref{results}. Results show an agreement between all evaluation measures on the boost obtained from multilingual pre-training. Accuracy increases of +1.3, +3.1 and +0.9 points are recorded using \textit{BLEU} \citep{papineni-etal-2002-bleu}, \textit{ChrF} \citep{popovic-2015-chrf} and \textit{Meteor} \citep{meteor} metrics respectively.
\vspace*{-2mm}
\begin{table}[t]
\caption{Automatic evaluation results for baseline approach and at each stages of our training pipeline.}.
\label{results}
\begin{center}
\begin{tabular}{l|ccc}
  & BLEU  & ChrF  & Meteor  \\ \hline \\
Baseline  & 22.28 &	46.51 &	26.1 \\
Multilingual   & 15.84 &	40.99 &	23.32  \\
Tigrinya  & 17.8 &	42.92 &	24.61 \\
In-domain & \textbf{23.6} &	\textbf{49.59} & \textbf{27.04} \\
\end{tabular}
\end{center}
\end{table}

\vspace*{-2mm}

\section{Conclusion}
\vspace*{-1mm}

With this work, we have demonstrated the utility of cross-lingual transfer learning on building a Tigrinya-to-English MT system. As a result of this work, a demonstration application was launched as the first neural Tigrinya-to-English translator\footnote{\url{http://gamayun.translatorswb.org/tigrinya}}. As for future work, we will develop English-to-Tigrinya models and evaluate the usability of the bidirectional system in a humanitarian setting using feedback from native speakers.

\subsubsection*{Acknowledgments}
This work was done partially in collaboration with the Masakhane initiative\footnote{\url{http://www.masakhane.io/}}. Special thanks to Musie Meressa Berhe for helping revise our dataset.

\bibliography{iclr2020_conference}
\bibliographystyle{iclr2020_conference}

\end{document}